\documentclass[letterpaper]{article} 
\usepackage{aaai2026}  
\usepackage{times}  
\usepackage{helvet}  
\usepackage{courier}  
\usepackage[hyphens]{url}  
\usepackage{graphicx} 
\usepackage{natbib}  
\usepackage{caption} 
\usepackage{algorithm}
\usepackage{algorithmic}
\usepackage[dvipsnames]{xcolor}
\usepackage[pagebackref=true,breaklinks=true,colorlinks=true,citecolor=NavyBlue,bookmarks=false]{hyperref}

\usepackage{newfloat}
\usepackage{listings}
\DeclareCaptionStyle{ruled}{labelfont=normalfont,labelsep=colon,strut=off} 
\floatstyle{ruled}
\newfloat{listing}{tb}{lst}{}
\floatname{listing}{Listing}
\usepackage{booktabs}
\usepackage{multirow}
\usepackage{adjustbox}
\usepackage{amsmath}
\usepackage{amsfonts}

\newcommand\eg{\textit{e.g.}}
\newcommand\ie{\textit{i.e.}}

\title{QuantVSR: Low-Bit Post-Training Quantization for \\ Real-World Video Super-Resolution}
\author{
    Bowen Chai$^{1}$\equalcontrib,\enspace
    Zheng Chen$^{1}$\equalcontrib,\enspace 
    Libo Zhu$^{1}$,\enspace\\
    Wenbo Li$^{2}$,\enspace
    Yong Guo$^{3}$,\enspace
    Yulun Zhang$^{1}$\thanks{Corresponding author: Yulun Zhang, yulun100@gmail.com}
}
\affiliations{
  \textsuperscript{1}Shanghai Jiao Tong University,\enspace\\
  \textsuperscript{2}Joy Future Academy,\enspace\\
  \textsuperscript{3}Max Planck Institute for Informatics
}

\begin{document}

\maketitle

\begin{abstract}
Diffusion models have shown superior performance in real-world video super-resolution (VSR). However, the slow processing speeds and heavy resource consumption of diffusion models hinder their practical application and deployment. Quantization offers a potential solution for compressing the VSR model. Nevertheless, quantizing VSR models is challenging due to their temporal characteristics and high fidelity requirements. To address these issues, we propose QuantVSR, a low-bit quantization model for real-world VSR. We propose a spatio-temporal complexity aware (STCA) mechanism, where we first utilize the calibration dataset to measure both spatial and temporal complexities for each layer. Based on these statistics, we allocate layer-specific ranks to the low-rank full-precision (FP) auxiliary branch. Subsequently, we jointly refine the FP and low-bit branches to achieve simultaneous optimization. In addition, we propose a learnable bias alignment (LBA) module to reduce the biased quantization errors. Extensive experiments on synthetic and real-world datasets demonstrate that our method obtains comparable performance with the FP model and significantly outperforms recent leading low-bit quantization methods. 
\end{abstract}

\setlength{\abovedisplayskip}{3pt}
\setlength{\belowdisplayskip}{3pt}

\begin{links}
    \link{Code}{https://github.com/bowenchai/QuantVSR}
\end{links}

\section{Introduction}
Video super-resolution (VSR) is a crucial task that aims to reconstruct high-resolution (HR) videos from low-resolution (LR) inputs. Early VSR methods~\cite{jo2018deep,wang2019edvr,nah2019ntire,chan2021basicvsr,liang2024vrt} assume synthetic degradations (\eg, bicubic). However, models trained on such degraded data perform poorly in restoring practical videos. In real-world scenarios, videos often suffer from various complex and unknown degradations, which increase the difficulty of video restoration. Numerous methods are proposed~\cite{yang2021real,pan2021deep,xie2023mitigating} to address this challenge. Among them, generative adversarial networks (GANs) gather attention for their ability to restore fine details~\cite{goodfellow2014generative,lucas2019generative}. Nevertheless, they still face challenges such as over-smoothing and training instability.

\begin{figure}[t]
\begin{center}
\scriptsize
\scalebox{1}{
    \hspace{-0.4cm}
    \begin{adjustbox}{valign=t}
    \begin{tabular}{cccc}
    \includegraphics[width=0.24\columnwidth]{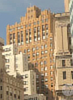} \hspace{-4mm} &
    \includegraphics[width=0.24\columnwidth]{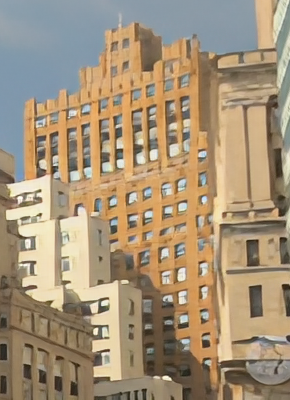} \hspace{-4mm} &
    \includegraphics[width=0.24\columnwidth]{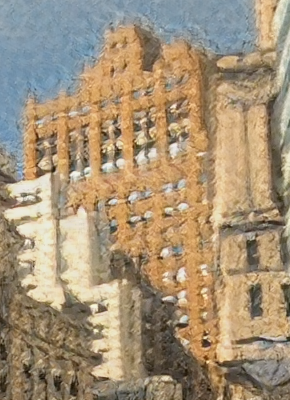} \hspace{-4mm} &
    \includegraphics[width=0.24\columnwidth]{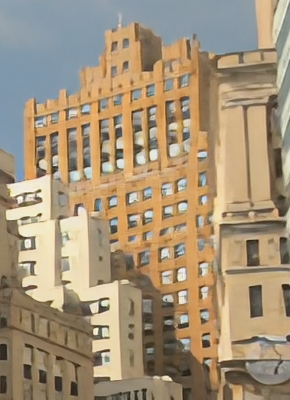} \hspace{-4mm} 
    \vspace{-0.5mm} \\
    LR ($\times$4) \hspace{-4mm} &
    MGLD-VSR \hspace{-4mm} &
    SVDQuant \hspace{-4mm} &
    QuantVSR (ours) \hspace{-4mm} \\
    Bits \hspace{-4mm} &
    W32A32 \hspace{-4mm} & 
    W4A4 \hspace{-4mm} &
    W4A4 \hspace{-5mm} \\
    
    \end{tabular}
    \end{adjustbox}
}
\end{center}
\vspace{-5mm}
\caption{Visual comparison among the full-precision VSR model, SVDQuant~\cite{li2024svdquant}, and our QuantVSR.}
\label{fig:page1_compare}
\vspace{-7mm}
\end{figure}

\begin{figure*}[t]
    \centering
    \scriptsize
    \includegraphics[width=\linewidth]{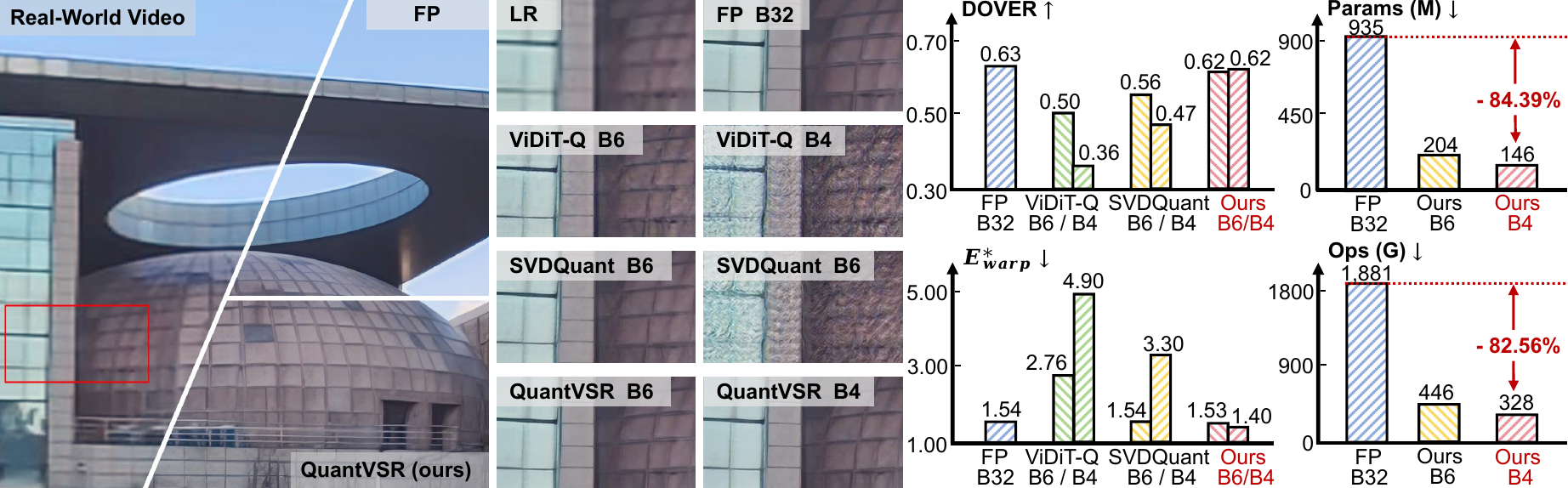}
    \vspace{-4mm}
    \caption{Performance comparisons on the real-world benchmark (\ie, MVSR4x~\cite{wang2023benchmark}) and compression ratio of our method. Bn represents n-bit quantization. Our QuantVSR compresses the model but retains performance comparable to the FP model, which surpasses existing quantization methods (\eg, ViDiT-Q~\cite{zhao2024vidit} and SVDQuant~\cite{li2024svdquant}). }
    \vspace{-4mm}
    \label{fig:intro-visual}
\end{figure*}

A new generative model, the diffusion model~\cite{ho2020denoising}, has recently shown promising results across various tasks, including VSR. Compared to GANs, diffusion models demonstrate superior generative performance with robust training stability. Diffusion-based VSR models~\cite{blattmann2023stable,yang2024motion,zhou2024upscale,xie2025star} leverage the generative capability while incorporating temporal modules, achieving remarkable visual quality. However, the high performance of diffusion models often comes with slow processing speeds and substantial resource consumption, which poses challenges for practical application and deployment on edge devices. 

To tackle these issues, quantization stands out as an effective compression method~\cite{nagel2021white,liu2025low}. Model quantization maps floating-point weights and/or activations to low-precision representations, typically as integers, thereby decreasing computational and memory costs. Prior studies have demonstrated the impressive results of quantization in image/video generation~\cite{li2023q,zhao2024vidit,li2024svdquant} and image restoration~\cite{zhu2025passionsr} tasks. However, in VSR, model quantization has rarely been explored, especially for the diffusion-based models. There are two critical difficulties in low-bit quantization for VSR models: \textbf{(1) Temporal consistency loss.} Model quantization introduces inconsistent errors across frames, resulting in a loss of temporal coherence in the final generated video. \textbf{(2) Complex data distribution.} VSR models embed temporal dynamics into latent features, resulting in more intricate activation distributions. This increases quantization challenges, as both spatial and temporal features must be considered to narrow the performance gap between full-precision (FP) and its quantized counterpart.

In this work, we propose QuantVSR, an effective low-bit quantization model for real-world VSR. We select MGLD-VSR~\cite{yang2024motion} as our FP backbone due to its outstanding performance. We adopt several common techniques from low-bit quantization, such as the high-precision branch~\cite{li2024svdquant} and rotation~\cite{ashkboos2024quarot}. Two novel designs are proposed to address the aforementioned issues. \textbf{First,} leveraging the unique temporal characteristics of VSR, we propose a spatio-temporal complexity aware (STCA) mechanism. A special low-rank branch is employed to skip the quantization process and is directly connected to the output. We allocate layer-specific ranks to the FP branch matrices based on the temporal and spatial complexities of the calibration dataset. Thus we achieve a trade-off between performance and computational efficiency. Subsequently, joint refinement of the FP and low-bit branches is applied to achieve simultaneous optimization. \textbf{Second,} we propose a learnable bias alignment (LBA) module to the quantization process, aiming to mitigate the severe biased quantization errors in low-bit quantization.

Comprehensive experiments (\ie, Figs. \ref{fig:page1_compare} and \ref{fig:intro-visual}) demonstrate that QuantVSR experiences minimal performance degradation even with 4-bit quantization. It outperforms recent leading quantization methods significantly. Compared to the FP model, 4-bit QuantVSR reduces parameter (Params) and operation (Ops) numbers by \textbf{84.39\%} and \textbf{82.56\%}, respectively (Fig.~\ref{fig:intro-visual}). Our contributions are:
\begin{itemize}
    \item We propose QuantVSR, a low-bit quantized model for real-world VSR. To the best of our knowledge, this is the first work to explore low-bit quantization (\eg, 4-bit and 6-bit) for diffusion-based VSR models. 
    \item Our spatio-temporal complexity aware (STCA) mechanism effectively preserves the performance of the FP model in both the temporal and spatial domains. Learnable bias alignment (LBA) module is proposed to reduce the severe biased error in low-bit quantization.
    \item Extensive experiments on the synthetic and real-world datasets demonstrate the superior performance of our method in comparison to other quantization methods.
\end{itemize}

\section{Related Work}
\subsection{Video Super-Resolution}
Video super-resolution (VSR) aims to recover high-resolution (HR) videos from low-resolution (LR) inputs. With the advancement of deep learning, various methods~\cite{jo2018deep,chan2021basicvsr,chan2022basicvsr++,liang2024vrt} have shown promising results, roughly categorized into recurrent-based~\cite{liang2022recurrent,shi2022rethinking} and sliding-window-based~\cite{yi2019progressive,li2020mucan} approaches. However, these methods often perform poorly on real-world videos due to their assumption of a fixed degradation process~\cite{liu2013bayesian,xue2019video}. A growing focus has emerged on real-world VSR recently, aiming to address complex and unknown degradations. RealVSR~\cite{yang2021real} and MVSR4x~\cite{wang2023benchmark} propose leveraging HR-LR paired data from real environments. While some approaches~\cite{chan2022investigating,xie2023mitigating} introduce diverse degradations for data augmentation to avoid labor-intensive data collection. Moreover, some works modify the model's structure to enhance its adaptability in real-world VSR, such as kernel estimation based on the image formation model~\cite{pan2021deep} and selective cross-attention modules~\cite{xie2023mitigating}. Although significant progress has been made, these methods still struggle to generate realistic details and fine textures.

\begin{figure*}[t]
    \centering
    \scriptsize
    \includegraphics[width=\linewidth]{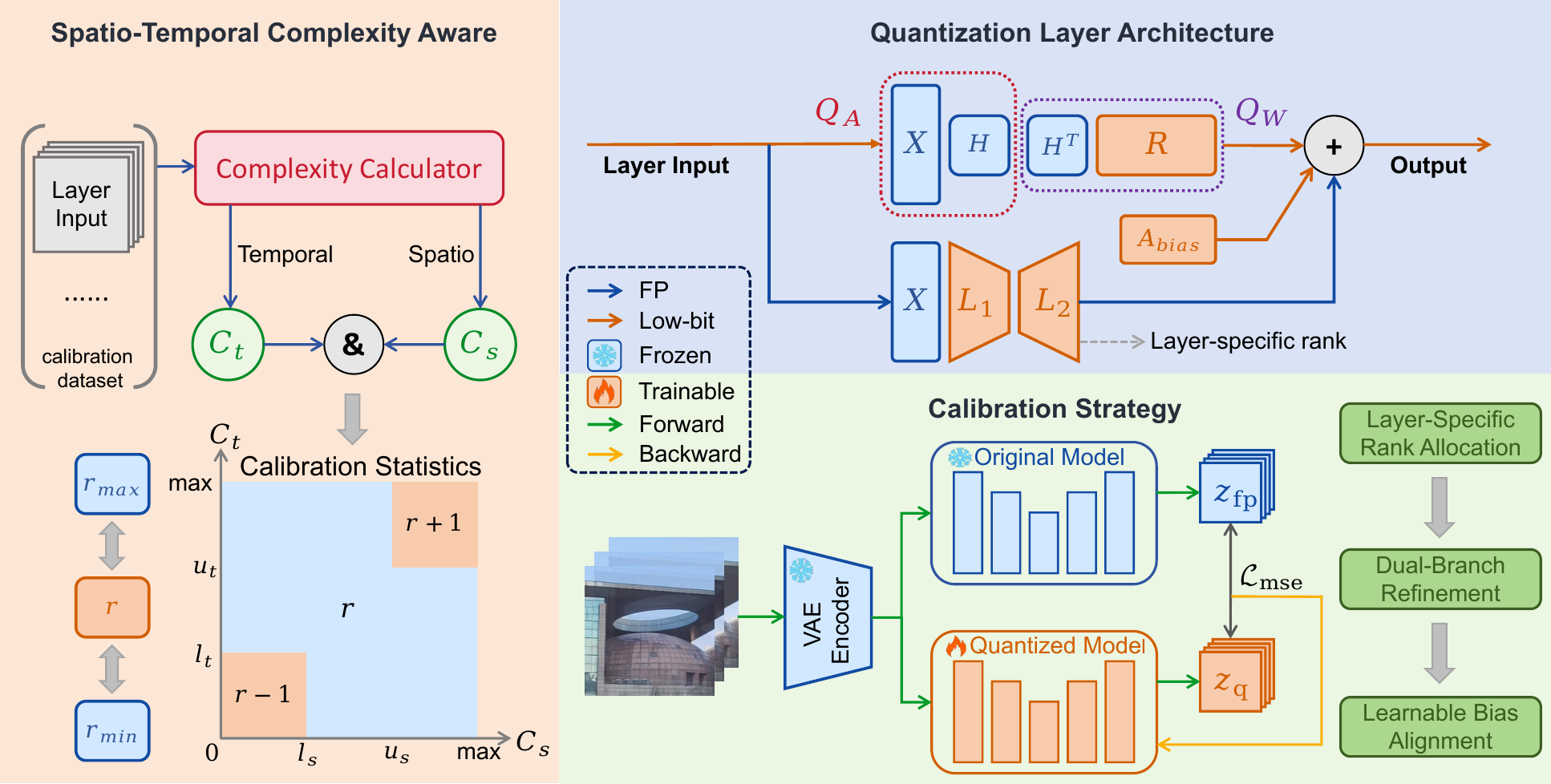}
    \vspace{-5mm}
    \caption{Overview of our QuantVSR. \textbf{First}, we analyze the temporal and spatial complexity distribution of the calibration dataset and leverage these statistics to allocate layer-specific ranks. \textbf{Next}, we jointly refine the two branches in spatio-temporal complexity aware mechanism. \textbf{Finally}, we train the learnable bias alignment module.}
    \vspace{-5mm}
    \label{fig:framework}
\end{figure*}

\subsection{Diffusion Model}
Diffusion models~\cite{ho2020denoising} demonstrate remarkable performance initially in image generation~\cite{ramesh2022hierarchical,rombach2022high} and have been extended to diverse vision tasks, including image restoration~\cite{wang2023zero,fei2023generative}, video generation~\cite{blattmann2023stable,chen2024videocrafter2,zhang2023i2vgen}, and video super-resolution (VSR)~\cite{zhou2024upscale,yang2024motion,he2024venhancer,li2025diffvsr,xie2025star,wang2025seedvr}. With rich generative priors encoded in pretrained diffusion models, diffusion-based approaches can generate highly realistic details, driving a new wave of VSR methods. Among them, some approaches extend pretrained image generation models~\cite{rombach2022high} with temporal mechanisms, such as adding temporal layers~\cite{zhou2024upscale} or guiding the denoising process with optical flows of LR videos~\cite{yang2024motion}. Other methods~\cite{he2024venhancer,xie2025star} directly leverage video generation models~\cite{zhang2023i2vgen}. Despite their success, diffusion-based VSR models are often associated with high computational and memory costs.

\subsection{Model Quantization}
Quantization is an effective method for model compression, as evidenced by its widespread and successful applications in the field of large language models~\cite{hu2022lora,liu2023llm,xiao2023smoothquant,shao2023omniquant,huang2024empirical}. With the rapid evolution of diffusion models, researchers have concentrated on enhancing their efficiency by applying quantization techniques. PTQ4DM~\cite{shang2023ptq4dm} pioneers the study of quantized diffusion models, achieving 8-bit quantization. Subsequent works have made breakthroughs with methods such as specialized calibration strategies~\cite{li2023q,ptqd}, sensitivity analysis~\cite{yang2023efficient}, quantization-aware training~\cite{li2023qdm}, weight-only quantization~\cite{sui2024bitsfusion}, and outlier removal~\cite{zhao2024vidit}. Recently, low-bit quantization has made significant progress. EfficientDM~\cite{he2023efficientdm} proposes a low-rank quantization fine-tuning strategy. While SVDQuant~\cite{li2024svdquant} employs a 16-bit parallel low-rank branch. Both EfficientDM and SVDQuant achieve 4-bit quantization with minimal performance loss. Furthermore, there are methods designed specifically for particular tasks and models. such as PassionSR~\cite{zhu2025passionsr} for one-step diffusion-based super-resolution models, ViDiT-Q~\cite{zhao2024vidit} for image and video generation. However, studies on the quantization of VSR models remain limited. Existing quantization methods cause significant performance loss when applied to VSR models.

\section{Method}
In this section, we introduce our low-bit quantized real-world VSR model, QuantVSR. First, we explain some necessary quantization preliminaries. Then, we present the overall framework of QuantVSR and the calibration process. Finally, we focus on two key designs proposed to mitigate the accuracy loss of quantizing VSR model: the spatio-temporal complexity aware (STCA) mechanism and the learnable bias alignment (LBA) module.
\subsection{Quantization Preliminary}
Model quantization reduces storage and computational costs by mapping weights and/or activations to low‑bit integers. The most common quantization process can be defined as:
\begin{equation}
    x_{\text{int}} = \text{Clip}\left(\left\lfloor\frac{x}{s}\right\rceil-z, l, u\right), \hat{x}=s\cdot x_{\text{int}}+z,
\end{equation}
where $x_{\text{int}}$ refers to the quantized integer and $\hat{x}$ is the de-quantized floating-point value. $s$ and $z$ denote the scale factor and zero point, which are related to data distribution and quantization bit‑width. $\lfloor\cdot\rceil$ is the round-to-nearest operator. $\text{Clip}(\cdot,l,u)$ ensures values remain within the range $[l, u]$.

Quantization‑aware training (QAT) and several recent post‑training quantization (PTQ)  methods rely on gradient back‑propagation, while the rounding operation introduced by quantization is non‑differentiable. To address this issue, the straight‑through estimator (STE) is widely adopted to approximate the gradients:
\begin{equation}
    \frac{\partial Q(x)}{\partial x} \approx
    \begin{cases}
    1 & \text{if } x \in [l,u],  \\
    0 & \text{otherwise}.
    \end{cases}
\end{equation}

\subsection{Overall Framework}
QuantVSR is built on MGLD-VSR~\cite{yang2024motion} for its superior performance in real-world video super-resolution (VSR). In the whole restoration process, the multi-step denoising of the diffusion model (\ie, UNet) dominates the computational cost compared to the VAE decoding stage. Therefore, we focus primarily on quantizing the UNet structure, by replacing the original layers (\ie, Linear, Conv2d, and Conv3d) with custom-designed quantized counterparts.

The overall architecture of the quantization layer is illustrated in Fig. \ref{fig:framework}. Spatio-temporal complexity aware (STCA) mechanism proposes an enhanced skip connection (SC) structure, where a full-precision (FP) branch allows part of the computation to bypass quantization.
For the low-bit branch, following previous work~\cite{ashkboos2024quarot}, layer inputs and weights are first smoothed using a randomized Hadamard transform to avoid outliers dominating limited quantization levels. Subsequently, the main computations are performed using low-bit integers. Learnable bias alignment (LBA) is added to the result finally. The entire quantization layer is formulated as follows:
\begin{equation}
    \boldsymbol{XW} = \underbrace{\boldsymbol{X}\boldsymbol{L_1L_2}}_{\text{FP, STCA}} + \underbrace{Q_A(\boldsymbol{XH})Q_W(\boldsymbol{H}^\top \boldsymbol{R})}_{\text{Low-Bit}} + \underbrace{\boldsymbol{A}_\text{bias}}_{\text{LBA}},
    \label{eq:quantlayer}
\end{equation}
where $L_1 \in \mathbb{R}^{m \times r}$ and $L_2 \in \mathbb{R}^{r \times n}$ are two low-rank matrices. $\boldsymbol{R} = \boldsymbol{W} - \boldsymbol{L_1L_2}$ is the residual. $\boldsymbol{H}$ is the random Hadamard matrix. $Q_W$ and $Q_A$ refer to the weight and activation quantizer. $\boldsymbol{A}_\text{bias}$ is the LBA module with the same dimensionality as the layer's bias.

The quantization calibration process consists of three main stages. First, we analyze the spatio-temporal complexity distribution of the calibration dataset. Such calibration statistics allocate higher ranks to the FP branches in layers with high complexity. Second, we refine two low-rank matrices and residual, allowing the STCA structure to attain optimal performance across both FP and low-bit branches in each layer. Finally, as changes in quantized weights introduced by previous modules affect the biased quantization errors, the LBA module is trained in the last stage with all other parameters frozen. For fine-tuning and training in the final two stages, we adopt the mean squared error (MSE) between FP and quantized model outputs (\ie, $\boldsymbol{z}_{fp}$ and $\boldsymbol{z}_{q}$).

\subsection{Spatio-Temporal Complexity Aware}
Under extremely low-bit quantization settings, the broad dynamic range of floating-point weights and activations is only represented by a limited set of integer values. This severely compromises the restoration ability of the quantized model. A commonly adopted strategy to mitigate this issue incorporates a low-rank full-precision (FP) branch to preserve the original model’s performance~\cite{li2024svdquant,he2023efficientdm}. However, this approach still has two main limitations: (1) The rank allocation strategy is suboptimal, incurring unnecessary overhead. (2) The FP branch may degrade the performance of the low-bit branch. Based on these analyses, we propose STCA that leverages the characteristics of video inputs in VSR model, consisting of two key steps: layer-specific rank allocation and dual-branch refinement.

\noindent\textbf{Layer-Specific Rank Allocation.} The FP branch in Eq.~\eqref{eq:quantlayer} maintains baseline accuracy by constraining the product $L_1L_2$ to approximate the original weight. A larger rank $r$ theoretically preserves more information, but it introduces a computational cost of $r \frac{m+n}{mn}$ that grows linearly with $r$. Consequently, gains in accuracy must be balanced against the accompanying increase in complexity.

A straightforward approach is to allocate layer‑specific ranks. Layers that process more complex inputs encode richer information and therefore have a greater impact on the final output. Accordingly, we assign them higher ranks to preserve more of the original performance. For VSR models, this complexity spans both temporal and spatial dimensions. Specifically, for the layer input $\boldsymbol{X} \in \mathbb{R}^{T \times C \times H \times W}$ ($T$: frames, $C$: channels, $H$: height, and $W$: width), we define its temporal complexity $C_{t}$ as:
\begin{equation}
    C_{t} = \frac{1}{T-1} \sum_{t=1}^{T-1} \frac{1}{CHW}\|\boldsymbol{X}_{t+1}-\boldsymbol{X}_t\|_2^2,
\end{equation}
where $\|\cdot\|_2$ denotes the Frobenius norm. This measures the inter-frame difference energy. Higher values reflect more intense motion between frames, which increases difficulty in restoration. Meanwhile, the spatial complexity is defined as:
\begin{equation}
    C_{s} = \frac{1}{TC}\sum_{t=1}^T\sum_{c=1}^C \sigma_{h,w}(\boldsymbol{X}_{t,c}),
\end{equation}
where $\sigma(\cdot)$ denotes the standard deviation calculated over the spatial dimensions (\ie, height and width). Since each channel represents a latent feature, its spatial variance reflects the complexity of that feature within each frame (\eg, textures or edges). Features with higher spatial variance are generally more informative and, therefore, should be assigned higher ranks to preserve overall performance.

As shown in Fig.~\ref{fig:framework}, the rank allocation process involves the following steps. First, we compute the distributions of temporal and spatial complexity for each layer using the calibration set, and define upper and lower thresholds for both (\ie,  $u_s$, $u_t$, $l_s$, and $l_t$). Then, for each layer input in the calibration set, we assess its complexity: (1) If both temporal and spatial complexities exceed the upper threshold, the rank is incremented by one. (2) If both fall below the lower threshold, the rank is decremented by one. (3) Otherwise, the rank remains unchanged. Finally, the rank is constrained to the interval $\left[r_{min}, r_{max}\right]$. To ensure compatibility with GPU‑parallel computation, the rank is then rounded to the nearest integer that is a multiple of $8$.

\begin{table*}[t]
    \centering
    \scriptsize
    \setlength{\tabcolsep}{2.1mm}
    \begin{tabular}{c | c | c c | c c c c c c c c c c}
        \toprule
        Dataset & Bits & SC & LBA & PSNR$\uparrow$ & SSIM$\uparrow$ & LPIPS$\downarrow$ & DISTS$\downarrow$ & CLIP‑IQA$\uparrow$ & MUSIQ$\uparrow$ & NIQE$\downarrow$ & MANIQA$\uparrow$ & DOVER$\uparrow$ & $E^*_{warp}\downarrow$\\
        \midrule
        \multirow{6}{*}{SPMCS} & \multirow{6}{*}{W4A4}
         & -- & -- & 17.13 & 0.2595 & 0.6480 & 0.4111 & 0.2553 & 44.14 & 7.26 & 0.2957 & 0.0802 & 10.00 \\
         & & -- & $\checkmark$ & 21.38 & 0.4996 & 0.3666 & 0.2352 & \underline{0.5097} & 64.74 & \textbf{3.37} & \underline{0.3183} & 0.6515 & 3.27 \\
         & & SVDSC & -- & 18.94 & 0.2820 & 0.5921 & 0.3756 & \textbf{0.5378} & 58.41 & 5.34 & \textbf{0.3471} & 0.4028 & 6.40 \\
         & & SVDSC & $\checkmark$ & 22.58 & 0.5783 & 0.3296 & 0.2089 & 0.4647 & 63.32 & 3.55 & 0.3144 & 0.6673 & 1.90 \\
         & & STCA & -- & \underline{22.75} & \underline{0.6071} & \underline{0.2914} & \underline{0.1816} & 0.4586 & \underline{65.49} & 3.45 & 0.3155 & \underline{0.6886} & \textbf{1.74} \\
         & & STCA & $\checkmark$ & \textbf{22.76} & \textbf{0.6075} & \textbf{0.2857} & \textbf{0.1747} & 0.4553 & \textbf{65.75} & \underline{3.41} & 0.3168 & \textbf{0.6969} & \underline{1.76} \\
        \bottomrule
    \end{tabular}
    \vspace{-1mm}
    \caption{Ablation study on our key designs: STCA and LBA. Experiments are conducted on the synthetic dataset SPMCS~\cite{yi2019progressive} with W4A4 quantization. SC stands for the skip connection structure. SVDSC means that the skip connection structure follows the same settings as SVDQuant~\cite{li2024svdquant}.}
    \label{tab:ablation}
    \vspace{-4mm}
\end{table*}

\noindent\textbf{Dual-Branch Refinement.} The FP branch preserves the original model's capabilities but alters the data distribution of the low-bit branch (\ie, $R$ in Eq. \eqref{eq:quantlayer}), affecting quantization difficulty. However, both branches contribute to the final output. This implies that enhancing the FP branch (\eg, by allocating a higher rank) may degrade the performance of the low-bit branch, leading to suboptimal overall results.

To ensure the effectiveness of both branches, we train two low-rank matrices $L_1$ and $L_2$ after layer-specific rank allocation, subject to the constraint $R = W-L_1L_2$. However, training them from scratch converges slowly, which undermines the goal of post-training quantization (PTQ). Inspired by SVDQuant~\cite{li2024svdquant}, we initialize the two matrices using singular value decomposition (SVD), providing a solid starting point for fine-tuning. With minimal training on the calibration set, the two branches achieve an optimal state, resulting in significantly improved performance.

\subsection{Learnable Bias Alignment}
The quantization error is biased, which means the average output of the FP model and its quantized counterpart is different. This issue is a major source of the overall quantization error, especially under low‑bit settings. Previous works have acknowledged this issue~\cite{nagel2019data,nagel2021white}, but have focused primarily on weight‑only quantization. They use calibration statistics or batch normalization (BN) parameters to reduce the discrepancy. For low-bit quantization of both weights and activations, the error becomes considerably more complex and can be formalized as follows:
\begin{equation}
    \mathbb{E}(\widehat{\boldsymbol{W}}\widehat{\boldsymbol{X}}) - \mathbb{E}(\boldsymbol{WX}) = \Delta\boldsymbol{W} \mathbb{E}(\widehat{\boldsymbol{X}}) + \boldsymbol{W} \mathbb{E}(\Delta\boldsymbol{X}),
    \label{eq:LBA}
\end{equation}
where $\widehat{\boldsymbol{W}} = \boldsymbol{W} + \Delta \boldsymbol{W}$ and $\widehat{\boldsymbol{X}}=\boldsymbol{X}+\Delta \boldsymbol{X}$ are quantized weights and activations. The weights and their quantization loss (\ie, $\boldsymbol{W}$ and $\Delta\boldsymbol{W}$) are fixed. So the error term is a function of the activation distribution and is complicated by activation quantization error (\ie, $\Delta \boldsymbol{X}$).

We propose a learnable bias alignment module to capture this error. As indicated by Eqs.  \eqref{eq:quantlayer} and \eqref{eq:LBA}, it is added to the output of the quantized layer and has the same dimensionality as the layer bias. The trainable parameters are relatively small compared with the whole model, allowing rapid convergence and preserving quantization efficiency. Moreover, during inference, this module can be fused into the layer bias, incurring negligible computational overhead.

\section{Experiments}

\subsection{Experimental Settings}

\noindent\textbf{Data Construction.} 
We sample intermediate input-output pairs from the full-precision (FP) UNet model on REDS30~\cite{nah2019ntire} at fixed intervals during the denoising process (\ie, 5 samples across 50 timesteps)~\cite{nagel2021white}. This results in a calibration set of 1,800 pairs, where each input has a shape of 5$\times$4$\times$64$\times$64. For evaluation, we apply both synthetic and real-world datasets. The synthetic datasets include REDS4~\cite{nah2019ntire} and SPMCS~\cite{yi2019progressive}, using multiple degradations (\ie, random blur, resizing, noise, JPEG compression, and video compression). For real-world datasets, we apply MVSR4x~\cite{wang2023benchmark}.

\begin{table}[t]
\centering
\scriptsize
\vspace{2mm}
\setlength{\tabcolsep}{3mm}
\begin{tabular}{c | c | c c}
\toprule
{Method} & {Bits} & {Params / M} ($\downarrow$ Ratio)  & {Ops / G} ($\downarrow$ Ratio)\\
\midrule
MGLD-VSR      & W32A32 & 935 ($\downarrow$0\%) & 1,881($\downarrow$0\%) \\
\midrule
\multirow{3}{*}{QuantVSR} & W8A8 & 263 ($\downarrow$71.87\%) & 563 ($\downarrow$70.07\%) \\
& W6A6 & 204 ($\downarrow$78.18\%) & 446 ($\downarrow$76.29\%)\\
& W4A4 & 146 ($\downarrow$84.39\%) & 328 ($\downarrow$82.56\%)\\
\bottomrule
\end{tabular}

\vspace{-1mm}
\caption{Params, Ops, and compression ratio (UNet only) of different quantization settings. Ops are computed with latent input size 5$\times$4$\times$64$\times$64, corresponding to a 5-frame video.}
\label{tab:Compression_ratio}
\vspace{-6mm}

\end{table}

\noindent\textbf{Evaluation Metrics.} We apply both image quality assessment (IQA) and video quality assessment (VQA). The IQA include reference-based metrics: PSNR, SSIM~\cite{wang2004image}, LPIPS~\cite{zhang2018unreasonable}, and DISTS~\cite{ding2020image}, and no-reference metrics: CLIP-IQA~\cite{wang2023exploring}, MUSIQ~\cite{ke2021musiq}, NIQE~\cite{zhang2015feature}, and MANIQA~\cite{yang2022maniqa}. For VQA, we adopt two metrics: DOVER~\cite{wu2023exploring} and $E^*_{warp}$~\cite{lai2018learning}.

\noindent\textbf{Implementation Details.} We quantize the weights and activations in the main components (\ie, UNet) with low bit-widths (\eg, 6 and 4 bits). $r_{min}$ and $r_{max}$ in STCA are set to 16 and 64 respectively. For both temporal and spatial complexity, the lower and upper thresholds (\ie, $l_s$, $l_t$, $u_s$, and $u_t$) are set to the 25th and 75th percentiles of the calibration dataset’s complexity distribution, respectively. The calibration training is performed on NVIDIA RTX A6000 GPU for 2 epochs. The learning rate is set as 1$\times$10$^{-3}$ and 2$\times$10$^{-4}$ during the first and second epoch, respectively.

\noindent\textbf{Compared Methods.} We select several representative and leading quantization methods: MaxMin~\cite{jacob2018quantization}, Q-Diffusion~\cite{li2023q}, QuaRot~\cite{ashkboos2024quarot}, ViDiT-Q~\cite{zhao2024vidit}, and SVDQuant~\cite{li2024svdquant}. We implement these methods on MGLD-VSR~\cite{yang2024motion} based on their released code.

\begin{table*}[t]
    \centering
    \scriptsize
    \setlength{\tabcolsep}{1.95mm}
    \begin{tabular}{c | c | c | c c c c c c c c c c}
        \toprule
        Datasets & Bits & Methods & PSNR$\uparrow$ & SSIM$\uparrow$ & LPIPS$\downarrow$ & DISTS$\downarrow$ & CLIP-IQA$\uparrow$
        & MUSIQ$\uparrow$ & NIQE$\downarrow$ & MANIQA$\uparrow$ & DOVER$\uparrow$ & $E^*_{warp}\downarrow$\\
        
        \midrule
        \multirow{13}{*}{REDS4} & W32A32 & MGLD-VSR & 23.27 & 0.6180 & 0.2117 & 0.0890 & 0.3641 & 65.81 & 2.65 & 0.3254 & 0.6761 & 7.24 \\
        \cline{2-13}
        & \multirow{6}{*}{W6A6} & MaxMin & 22.53 & 0.5059 & 0.4150 & 0.2444 & \textbf{0.3905} & 55.68 & 2.77 & 0.2512 & 0.6044 & 8.84 \\
        & & Q-Diffusion & 22.99 & 0.5589 & 0.3534 & 0.2036 & 0.3402 & 56.43 & \underline{2.62} & 0.2389 & 0.6244 & 7.40 \\
        & & QuaRot & \textbf{23.42} & \textbf{0.6213} & 0.2388 & \underline{0.1179} & 0.2731 & 60.66 & 2.80 & 0.2639 & 0.6562 & \textbf{6.66} \\
        & & ViDiT-Q & 22.68 & 0.5670 & 0.2759 & 0.1461 & 0.3346 & 62.86 & 2.66 & 0.2918 & 0.6267 & 8.75 \\
        & & SVDQuant & 23.26 & 0.6035 & \underline{0.2379} & 0.1211 & 0.3289 & \underline{63.89} & \textbf{2.54} & \underline{0.3018} & \underline{0.6717} & \underline{7.02} \\
        & & QuantVSR (ours) & \underline{23.30} & \underline{0.6167} & \textbf{0.2138} & \textbf{0.0921} & \underline{0.3614} & \textbf{65.46} & 2.63 & \textbf{0.3230} & \textbf{0.6755} & 7.12 \\
        \cline{2-13}
        & \multirow{6}{*}{W4A4} & MaxMin & 16.18 & 0.1995 & 0.6720 & 0.4026 & 0.2821 & 46.52 & 5.90 & 0.2931 & 0.1451 & 52.27 \\
        & & Q-Diffusion & 19.99 & 0.3176 & 0.5279 & 0.3096 & \textbf{0.5887} & 52.81 & 3.69 & \textbf{0.3415} & 0.4936 & 19.63 \\
        & & QuaRot & 20.21 & 0.3182 & 0.5346 & 0.3179 & \underline{0.5561} & 51.86 & 3.77 & \underline{0.3109} & 0.5195 & 17.82 \\
        & & ViDiT-Q & 20.98 & 0.4118 & \underline{0.4436} & \underline{0.2523} & 0.4503 & 52.85 & \underline{3.03} & 0.2986 & 0.5404 & 14.93 \\
        & & SVDQuant & \underline{21.19} & \underline{0.4138} & 0.4718 & 0.2908 & 0.5116 & \underline{54.09} & 3.18 & 0.2699 & \underline{0.5865} & \underline{12.46} \\
        & & QuantVSR (ours) & \textbf{23.31} & \textbf{0.6143} & \textbf{0.2286} & \textbf{0.1055} & 0.3364 & \textbf{64.19} & \textbf{2.61} & 0.3046 & \textbf{0.6822} & \textbf{6.88} \\

        \midrule
        \multirow{13}{*}{SPMCS} & W32A32 & MGLD-VSR & 22.81 & 0.6157 & 0.2807 & 0.1643 & 0.4500 & 65.89 & 3.56 & 0.3211 & 0.6957 & 1.81 \\
        \cline{2-13}
        & \multirow{6}{*}{W6A6} & MaxMin & 21.79 & 0.4891 & 0.4395 & 0.2914 & \textbf{0.5163} & 61.08 & 3.44 & 0.2997 & 0.5978 & 2.45 \\
        & & Q-Diffusion & 22.37 & 0.5504 & 0.3937 & 0.2618 & \underline{0.4788} & 60.19 & \textbf{3.41} & 0.2844 & 0.5912 & 1.77 \\
        & & QuaRot & \textbf{22.95} & \textbf{0.6173} & \underline{0.2991} & \underline{0.1892} & 0.3716 & 60.64 & 3.75 & 0.2839 & 0.6251 & \textbf{1.66} \\
        & & ViDiT-Q & 22.08 & 0.5652 & 0.3204 & 0.2158 & 0.4590 & 63.36 & \underline{3.41} & \underline{0.3106} & 0.6630 & 2.43 \\
        & & SVDQuant & 22.74 & 0.5959 & 0.3118 & 0.2013 & 0.4541 & \underline{63.48} & 3.51 & 0.3066 & \underline{0.6724} & \underline{1.76} \\
        & & QuantVSR (ours) & \underline{22.80} & \underline{0.6126} & \textbf{0.2833} & \textbf{0.1697} & 0.4569 & \textbf{65.73} & 3.54 & \textbf{0.3215} & \textbf{0.6848} & 1.79 \\
        \cline{2-13}
        & \multirow{6}{*}{W4A4} & MaxMin & 17.51 & 0.2819 & 0.6239 & 0.3886 & 0.2823 & 47.07 & 6.55 & 0.3040 & 0.1331 & 8.92 \\
        & & Q-Diffusion & 19.85 & 0.3370 & 0.5275 & 0.3381 & \textbf{0.5765} & \underline{60.48} & 4.67 & \textbf{0.3568} & 0.4996 & 4.82 \\
        & & QuaRot & 20.16 & 0.3505 & 0.5277 & 0.3397 & \underline{0.5691} & 59.44 & 4.52 & \underline{0.3246} & 0.5554 & 4.40 \\
        & & ViDiT-Q & 20.60 & 0.4276 & \underline{0.4600} & \underline{0.2911} & 0.4984 & 59.55 & 3.78 & 0.3075 & 0.5294 & 3.98 \\
        & & SVDQuant & \underline{21.17} & \underline{0.4459} & 0.4739 & 0.3126 & 0.5417 & 57.39 & \underline{3.74} & 0.2897 & \underline{0.5627} & \underline{2.77} \\
        & & QuantVSR (ours) & \textbf{22.76} & \textbf{0.6075} & \textbf{0.2857} & \textbf{0.1747} & 0.4553 & \textbf{65.75} & \textbf{3.41} & 0.3168 & \textbf{0.6969} & \textbf{1.76} \\

        \midrule
        \multirow{13}{*}{MVSR4x} & W32A32 & MGLD-VSR & 22.77 & 0.7422 & 0.3571 & 0.2248 & 0.3699 & 53.65 & 5.06 & 0.2880 & 0.6321 & 1.54 \\
        \cline{2-13}
        & \multirow{6}{*}{W6A6} & MaxMin           & 22.29 & 0.5664 & 0.5478 & 0.3335 & \textbf{0.5734} & 51.38 & \textbf{3.92} & \textbf{0.3166} & 0.4538 & 2.47 \\
        &                       & Q\text{-}Diffusion & 22.59 & 0.6377 & 0.5011 & 0.2997 & \underline{0.5486} & 50.41 & 3.93 & 0.3041 & 0.4353 & 1.80 \\
        &                       & QuaRot           & \textbf{22.85} & \textbf{0.7494} & \textbf{0.3540} & \underline{0.2333} & 0.3194 & 47.08 & 5.51 & 0.2531 & 0.5463 & \textbf{1.40} \\
        &                       & ViDiT\text{-}Q   & 22.16 & 0.6396 & 0.4487 & 0.2861 & 0.5131 & 51.88 & \underline{3.92} & 0.3022 & 0.5043 & 2.76 \\
        &                       & SVDQuant         & \underline{22.84} & 0.7107 & 0.3842 & 0.2528 & 0.4720 & \underline{53.24} & 4.12 & \underline{0.3075} & \underline{0.5557} & 1.54 \\
        &                       & QuantVSR (ours)  & 22.80 & \underline{0.7319} & \underline{0.3559} & \textbf{0.2276} & 0.4201 & \textbf{54.68} & 4.62 & 0.3009 & \textbf{0.6156} & \underline{1.53} \\
        \cline{2-13}
        & \multirow{6}{*}{W4A4} & MaxMin           & 18.70 & 0.2813 & 0.7245 & 0.4330 & 0.2818 & 33.29 & 5.90 & 0.3146 & 0.1627 & 8.97 \\
        &                       & Q\text{-}Diffusion & 21.10 & 0.3971 & 0.6263 & 0.3855 & 0.5973 & 47.53 & 4.52 & \textbf{0.3660} & 0.3362 & 5.33 \\
        &                       & QuaRot           & 21.00 & 0.3880 & 0.6253 & 0.3919 & \underline{0.6197} & 49.09 & 4.88 & \underline{0.3416} & 0.3795 & 5.31 \\
        &                       & ViDiT\text{-}Q   & 21.18 & 0.4471 & 0.5956 & \underline{0.3600} & 0.5833 & 48.09 & \textbf{4.09} & 0.3253 & 0.3618 & 4.90 \\
        &                       & SVDQuant         & \underline{21.70} & \underline{0.5021} & \underline{0.5780} & 0.3659 & \textbf{0.6394} & \underline{53.68} & \underline{4.24} & 0.3328 & \underline{0.4727} & \underline{3.30} \\
        &                       & QuantVSR (ours)  & \textbf{22.90} & \textbf{0.7367} & \textbf{0.3590} & \textbf{0.2309} & 0.4339 & \textbf{55.35} & 4.57 & 0.3017 & \textbf{0.6219} & \textbf{1.40} \\
        
        \bottomrule
    \end{tabular}
    \vspace{-2mm}
    \caption{Quantitative results on synthetic and real-world datasets. The full-precision backbone is MGLD-VSR~\cite{yang2024motion}. \textbf{Bold} and \underline{underline} represent the best and second best scores, respectively.}
    \label{tab:whole}
    \vspace{-3mm}
\end{table*}

\begin{figure*}[!t]
\scriptsize
\centering
\begin{tabular}{cccccccc}

\hspace{-0.48cm}
\begin{adjustbox}{valign=t}
\begin{tabular}{c}
\includegraphics[width=0.22\textwidth]{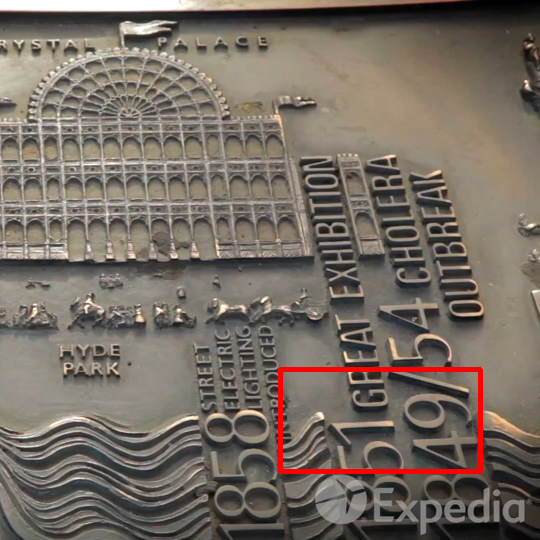}
\\
SPMCS: LDVTG\_009
\end{tabular}
\end{adjustbox}
\hspace{-0.46cm}
\begin{adjustbox}{valign=t}
\begin{tabular}{cccccc}
\includegraphics[width=0.189\textwidth]{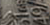} \hspace{-4.mm} &
\includegraphics[width=0.189\textwidth]{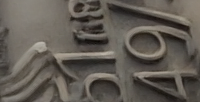} \hspace{-4.mm} &
\includegraphics[width=0.189\textwidth]{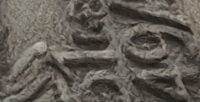} \hspace{-4.mm} &
\includegraphics[width=0.189\textwidth]{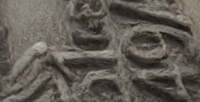} \hspace{-4.mm} &
\\ 
LR \hspace{-4.mm} &
MGLD-VSR / W32A32 \hspace{-4.mm} &
MaxMin / W6A6 \hspace{-4.mm} &
Q-Diffusion / W6A6 \hspace{-4.mm} &
\\
\includegraphics[width=0.189\textwidth]{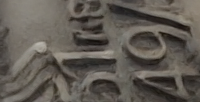} \hspace{-4.mm} &
\includegraphics[width=0.189\textwidth]{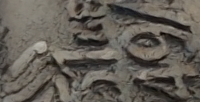} \hspace{-4.mm} &
\includegraphics[width=0.189\textwidth]{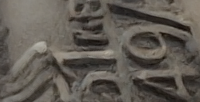} \hspace{-4.mm} &
\includegraphics[width=0.189\textwidth]{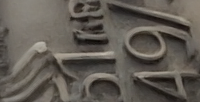} \hspace{-4.mm} &
\\ 
QuaRot / W6A6 \hspace{-4.mm} &
ViDiT-Q / W6A6 \hspace{-4.mm} &
SVDQuant / W6A6 \hspace{-4.mm} &
QuantVSR (ours) / W6A6 \hspace{-4mm}
\\
\end{tabular}
\end{adjustbox}
\\

\hspace{-0.48cm}
\begin{adjustbox}{valign=t}
\begin{tabular}{c}
\includegraphics[width=0.22\textwidth]{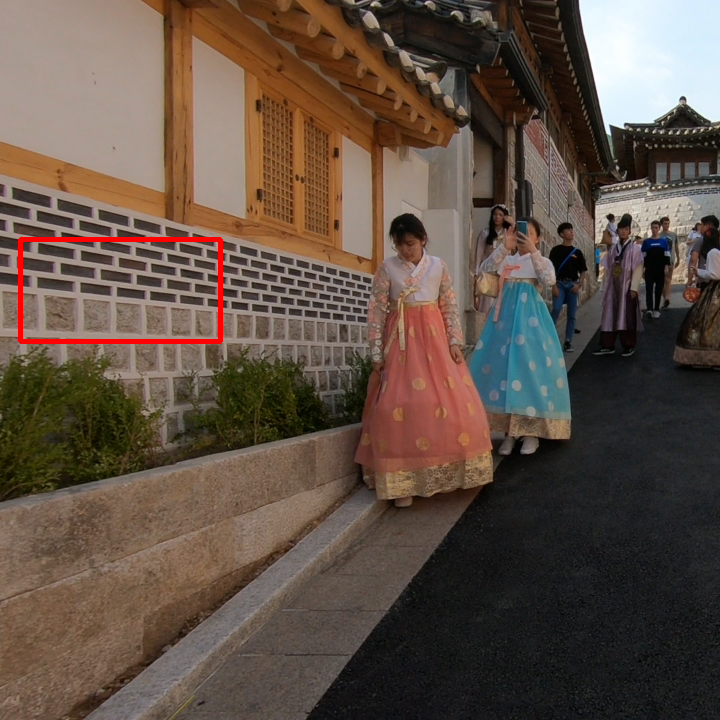}
\\
REDS4: 011
\end{tabular}
\end{adjustbox}
\hspace{-0.46cm}
\begin{adjustbox}{valign=t}
\begin{tabular}{cccccc}
\includegraphics[width=0.189\textwidth]{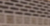} \hspace{-4.mm} &
\includegraphics[width=0.189\textwidth]{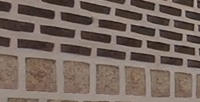} \hspace{-4.mm} &
\includegraphics[width=0.189\textwidth]{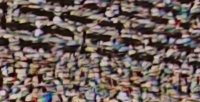} \hspace{-4.mm} &
\includegraphics[width=0.189\textwidth]{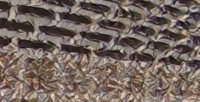} \hspace{-4.mm} &
\\ 
LR \hspace{-4.mm} &
MGLD-VSR / W32A32 \hspace{-4.mm} &
MaxMin / W4A4 \hspace{-4.mm} &
Q-Diffusion / W4A4 \hspace{-4.mm} &
\\
\includegraphics[width=0.189\textwidth]{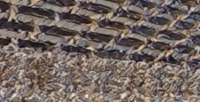} \hspace{-4.mm} &
\includegraphics[width=0.189\textwidth]{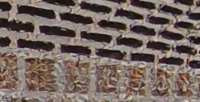} \hspace{-4.mm} &
\includegraphics[width=0.189\textwidth]{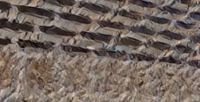} \hspace{-4.mm} &
\includegraphics[width=0.189\textwidth]{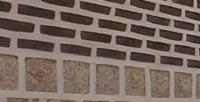} \hspace{-4.mm} &
\\ 
QuaRot / W4A4 \hspace{-4.mm} &
ViDiT-Q / W4A4 \hspace{-4.mm} &
SVDQuant / W4A4 \hspace{-4.mm} &
QuantVSR (ours) / W4A4 \hspace{-4mm}
\\
\end{tabular}
\end{adjustbox}
\\

\hspace{-0.48cm}
\begin{adjustbox}{valign=t}
\begin{tabular}{c}
\includegraphics[width=0.22\textwidth]{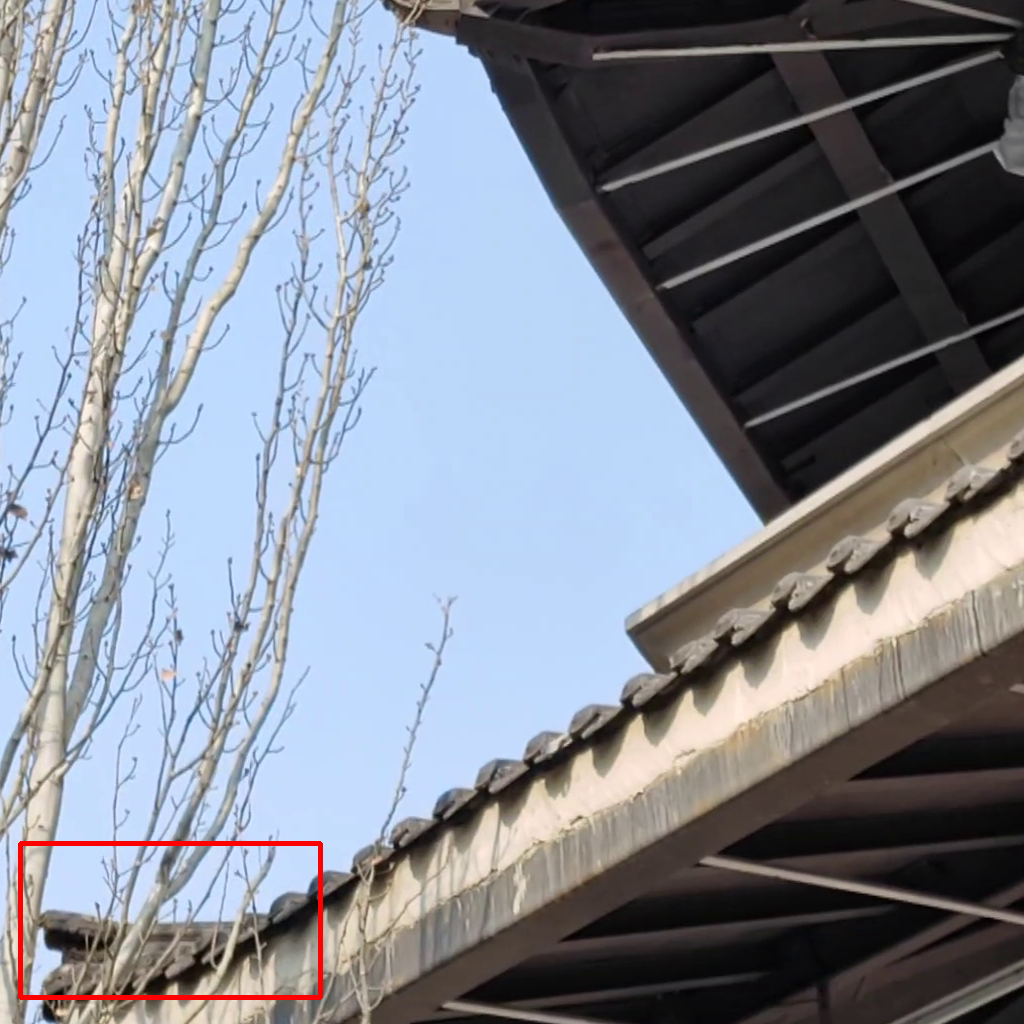}
\\
MVSR4x: 466
\end{tabular}
\end{adjustbox}
\hspace{-0.46cm}
\begin{adjustbox}{valign=t}
\begin{tabular}{cccccc}
\includegraphics[width=0.189\textwidth]{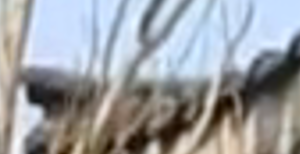} \hspace{-4.mm} &
\includegraphics[width=0.189\textwidth]{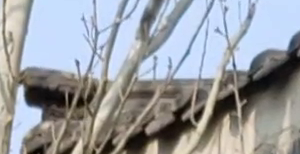} \hspace{-4.mm} &
\includegraphics[width=0.189\textwidth]{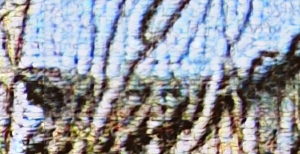} \hspace{-4.mm} &
\includegraphics[width=0.189\textwidth]{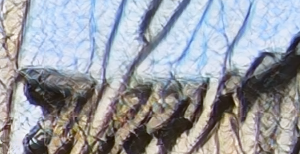} \hspace{-4.mm} &
\\ 
LR \hspace{-4.mm} &
MGLD-VSR / W32A32 \hspace{-4.mm} &
MaxMin / W4A4 \hspace{-4.mm} &
Q-Diffusion / W4A4 \hspace{-4.mm} &
\\
\includegraphics[width=0.189\textwidth]{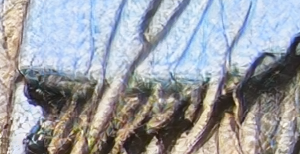} \hspace{-4.mm} &
\includegraphics[width=0.189\textwidth]{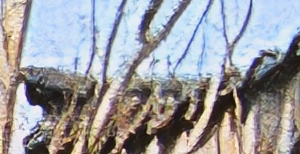} \hspace{-4.mm} &
\includegraphics[width=0.189\textwidth]{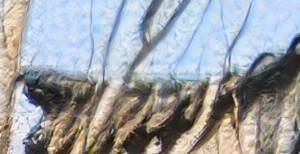} \hspace{-4.mm} &
\includegraphics[width=0.189\textwidth]{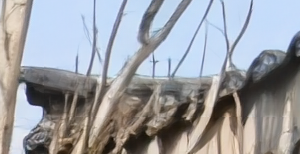} \hspace{-4.mm} &
\\ 
QuaRot / W4A4 \hspace{-4.mm} &
ViDiT-Q / W4A4 \hspace{-4.mm} &
SVDQuant / W4A4 \hspace{-4.mm} &
QuantVSR (ours) / W4A4 \hspace{-4mm}
\\
\end{tabular}
\end{adjustbox}
\\

\end{tabular}
\vspace{-3.mm}
\caption{Visual comparison on synthetic (SPMCS~\cite{yi2019progressive}, REDS4~\cite{nah2019ntire}) and real-world (MVSR4x~\cite{wang2023benchmark}) datasets at 6 / 4-bit quantization. Our approach outperforms existing methods especially in the 4‑bit setting.}
\label{fig:visual}
\vspace{-4.mm}
\end{figure*}

\subsection{Ablation Study}
\noindent\textbf{Spatio-Temporal Complexity Aware (STCA).} To verify the effectiveness of STCA, we conduct experiments under three configurations: (1) without the skip connection (SC) structure, (2) using the SVDQuant~\cite{li2024svdquant} settings (\ie, fixed rank and SVD, denoted as SVDSC), and (3) our proposed STCA method. As shown in Tab.~\ref{tab:ablation}, STCA demonstrates a significant performance improvement compared to SVDSC and the method without SC. 

\noindent\textbf{Learnable Bias Alignment (LBA).} We adopt LBA under various settings, as shown in Tab. \ref{tab:ablation}. In all cases, LBA leads to substantial improvements, particularly when accuracy degradation is severe. This strongly supports our previous conclusion regarding the importance of biased quantization errors in low-bit quantization scenarios. Although modest, integrating LBA after STCA still yields slight enhancements across multiple metrics.

\subsection{Main Results}

\noindent\textbf{Quantitative Results.} Quantitative comparisons are shown in Tab. \ref{tab:whole}. For VQA and reference-based IQA, our method outperforms prior methods on most datasets, especially in 4-bit settings. Some methods achieve high scores on no-reference IQA metrics (\eg, CLIP-IQA), even surpassing FP models by a large margin, yet their performance on structural metrics (\eg, PSNR and SSIM) remains poor. Prior studies~\cite{zhu2025passionsr} note that images with significant noise may still yield high no-reference scores. These approaches do not surpass ours, as further illustrated in Fig.~\ref{fig:visual}.

\begin{figure}[t]
    \centering
    \scriptsize
    \includegraphics[width=\linewidth]{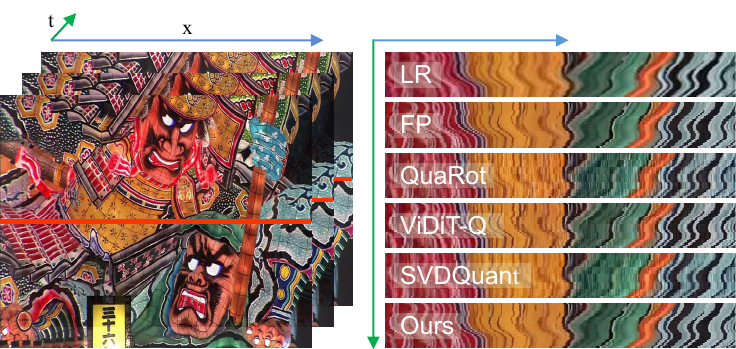}
    \vspace{-4mm}
    \caption{Comparison of temporal consistency (stacking the red line across frames).}
    \vspace{-5mm}
    \label{fig:temporal}
\end{figure}

\noindent\textbf{Qualitative Results.}
Qualitative comparisons are shown in Figs. \ref{fig:visual} and \ref{fig:temporal}. QuantVSR produces sharper details and more faithful textures than prior methods, with minimal difference from the FP model. Leading quantization methods tend to introduce unrealistic artifacts when applied to MGLD-VSR (\eg, ViDiT-Q on REDS4 011), whereas our QuantVSR suppresses such distortions. Moreover, our QuantVSR achieves higher temporal consistency than competing techniques, owing to our special consideration of temporal features.

\noindent\textbf{Rank in Full-Precision Branch.} The rank of two matrices in the skip connection structure determines the computational burden. Using our proposed STCA, most layers preserve the minimal rank $r_{min}$ (\ie, 16), and only a few complex layers require higher ranks (up to $r_{max}$, 64). Consequently, the average rank across all layers of QuantVSR is 24, which is lower than the fixed rank of 32 in SVDQuant~\cite{li2024svdquant} at 4-bit quantization.

\noindent\textbf{Compression Ratio.}
We calculate the model size (Params / M) and the number of operations (Ops / G) with the same method as previous quantization studies~\cite{qin2023quantsr}. Table \ref{tab:Compression_ratio} shows that 6‑bit QuantVSR shrinks model size and computation by 78.18\% and 76.29\%, while 4‑bit cuts them by 84.39\% and 82.56\% versus the FP MGLD-VSR.

\section{Conclusion}
In this paper, we propose QuantVSR, a low-bit quantization framework for real-world video super-resolution (VSR). To maintain full-precision (FP) model performance across both spatial and temporal dimensions, we propose a spatio-temporal complexity aware (STCA) mechanism. This also enables the joint optimization of both branches in the skip connection (SC) structure.  In addition, a learnable bias alignment (LBA) module mitigates the severe biased quantization error without incurring any additional overhead during inference. Comprehensive experiments on synthetic and real-world benchmarks demonstrate that QuantVSR achieves perceptual quality comparable to its FP counterpart at 6-bit and even 4-bit precision. It significantly outperforms the latest state-of-the-art quantization methods overall.

\section{Acknowledgments}
This work was supported by Shanghai Municipal Science and Technology Major Project (2021SHZDZX0102) and the Fundamental Research Funds for the Central Universities.

\end{document}